\pdfoutput=1

\documentclass[11pt]{article}

\usepackage[]{acl}
\usepackage{times}
\usepackage{latexsym}
\usepackage{booktabs}

\usepackage[T1]{fontenc}

\usepackage[utf8]{inputenc}

\usepackage{microtype}
\usepackage{amsmath}
\usepackage{amssymb}
\usepackage{bm}
\usepackage{todonotes}
\usepackage{comment}
\usepackage{tabularx}
\usepackage{multirow}
\usepackage{xcolor,colortbl}
\definecolor{Gray}{gray}{0.85}
\newcolumntype{g}{>{\columncolor{Gray}}c}
\usepackage{graphicx}
\makeatletter
\def\thickhline{%
  \noalign{\ifnum0=`}\fi\hrule \@height \thickarrayrulewidth \futurelet
   \reserved@a\@xthickhline}
\def\@xthickhline{\ifx\reserved@a\thickhline
               \vskip\doublerulesep
               \vskip-\thickarrayrulewidth
             \fi
      \ifnum0=`{\fi}}
\makeatother
\newlength{\thickarrayrulewidth}
\usepackage{color,soul}
\title{DS-TOD: Efficient Domain Specialization for Task-Oriented Dialog}

\author{Chia-Chien Hung\textsuperscript{1}, Anne Lauscher\textsuperscript{2}, Simone Paolo Ponzetto\textsuperscript{1}, and Goran Glava\v{s}\textsuperscript{1,3} \\
  \textsuperscript{1}Data and Web Science Group, University of Mannheim, Germany \\
  \textsuperscript{2}MilaNLP, Bocconi University, Italy \\
  \textsuperscript{3}Center for Information and Language Processing, LMU Munich, Germany \\ \texttt{\{chia-chien, simone, goran\}@informatik.uni-mannheim.de} \\ \texttt{anne.lauscher@unibocconi.it} \\}

\begin{document}

\maketitle

\begin{abstract}
Recent work has shown that self-supervised dialog-specific pretraining on large conversational datasets yields substantial gains over traditional language modeling (LM) pretraining in downstream task-oriented dialog (TOD). 
These approaches, however, exploit general dialogic corpora (e.g., Reddit) and thus presumably fail to reliably embed domain-specific knowledge useful for concrete downstream TOD domains.
In this work, we investigate the effects of domain specialization of pretrained language models (PLMs) for TOD. 
Within our DS-TOD framework, we first automatically extract salient domain-specific terms, and then use them to construct \textsc{DomainCC} and \textsc{DomainReddit} -- resources that we leverage for domain-specific pretraining, based on (i) masked language modeling (MLM) and (ii) response selection (RS) objectives, respectively. 
We further propose a resource-efficient and modular domain specialization by means of \textit{domain adapters} -- additional parameter-light layers in which we encode the domain knowledge. Our experiments with prominent TOD tasks -- dialog state tracking (DST) and response retrieval (RR) -- encompassing five domains from the \textsc{MultiWOZ} benchmark demonstrate the effectiveness of DS-TOD. Moreover, we show that the light-weight adapter-based specialization (1) 
performs comparably to full fine-tuning in single domain setups and (2) is particularly suitable
for multi-domain specialization, where besides advantageous computational footprint, it
can offer better TOD performance.

\end{abstract}

\section{Introduction}
Task-oriented dialog (TOD), where conversational agents help users complete concrete tasks (e.g., book flights or order food), has arguably been one of the most prominent NLP applications in recent years, both in academia \citep[\textit{inter alia}]{budzianowski-etal-2018-multiwoz, henderson-etal-2019-training, liu2021robustness} and industry~\citep[e.g.,][]{yan2017building, henderson-etal-2019-polyresponse}.
%
%
Like for most other NLP tasks, fine-tuning of pretrained language models (PLMs) like BERT~\citep{devlin-etal-2019-bert} and GPT-2~\citep{radford2019language} pushed the state-of-the-art in TOD tasks \citep{budzianowski-vulic-2019-hello, hosseini2020simple}, with LM pretraining at the same time alleviating the need for large labeled datasets~\citep{ramadan-etal-2018-large}. 


More recent TOD work recognized the idiosyncrasy of dialog -- i.e., that dialogs represent interleaved exchanges of utterances between two (or more) participants -- and proposed pretraining objectives specifically tailored for dialogic corpora \cite[\textit{inter alia}]{henderson-etal-2019-training, wu-etal-2020-tod, bao-etal-2020-plato}. For instance, \newcite{wu-etal-2020-tod} pretrain their TOD-BERT model on the concatenation of nine human-to-human multi-turn dialog datasets. Similarly, \citet{henderson-etal-2019-training,henderson-etal-2020-convert} pretrain a general-purpose dialog encoder on a large corpus from Reddit by means of response selection objectives. Encoding dialogic linguistic knowledge in this way led to significant performance improvements in downstream TOD tasks. 

\begin{figure*}[th]
	\centering
    \includegraphics[trim={1.05cm 0 1cm 0},clip,width=\textwidth]{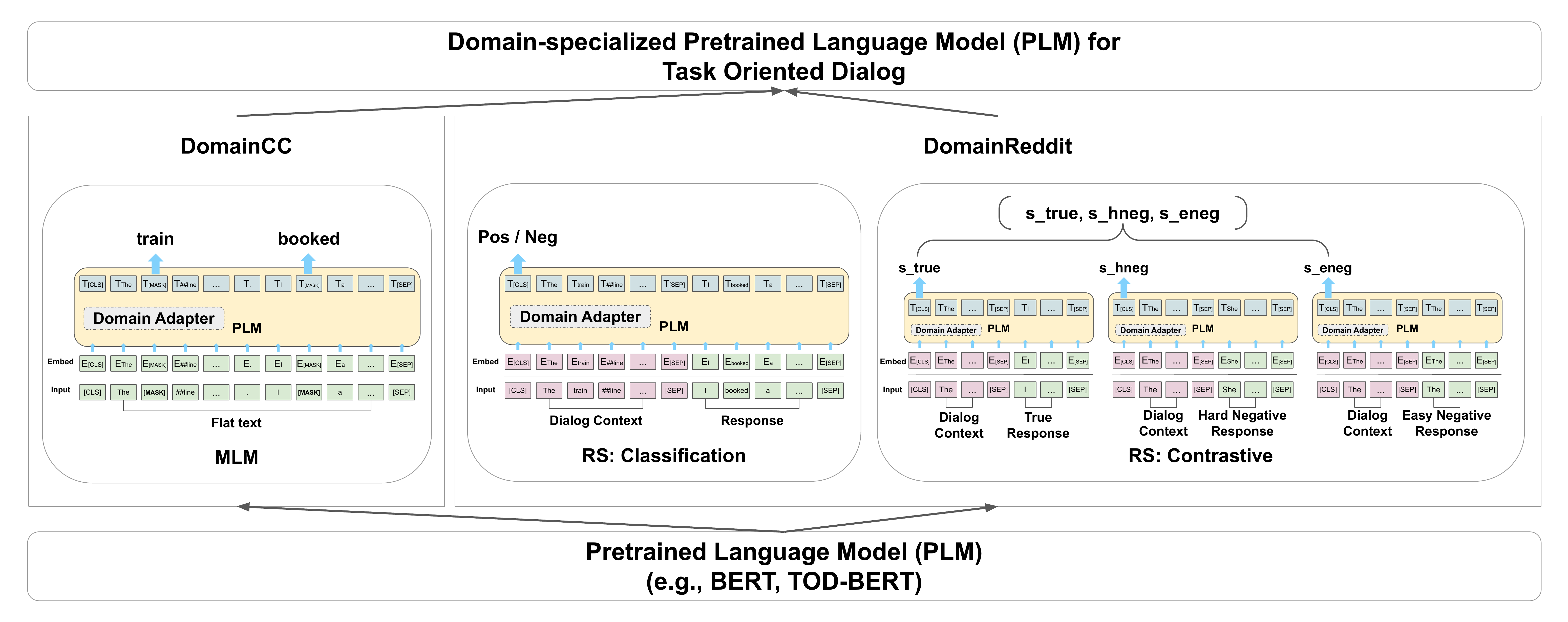}
	\caption{Overview of DS-TOD. Three different specialization objectives for injecting domain-specific knowledge into PLMs (see \S\ref{ss:objectives}): (1) Masked Language Modeling (MLM) on the ``flat'' domain corpus \textsc{DomainCC}, (2) Response Selection (RS) via Classification, and (3) Response Selection via Contrastive Learning operating on the dialogic \textsc{DomainReddit}. Domain specialization performed either via (a) full fine-tuning or (b) adapters (see \S\ref{ss:adapters}).} 
	\label{fig:DS-TOD}
\end{figure*}

While these approaches impart useful dialogic linguistic knowledge they fail to exploit the fact that individual task-oriented dialogs typically belong to one narrow domain (e.g., \textit{food} ordering) or few closely related domains~\citep[e.g., booking a \textit{train} and \textit{hotel};][]{budzianowski-etal-2018-multiwoz, ramadan-etal-2018-large}. 
Given the multitude of different downstream TOD domains (e.g., ordering \textit{food} is quite different from booking a \textit{flight}) it is, intuitively, unlikely that general dialogic pretraining reliably encodes domain-specific knowledge for all of them.   
In this work, we propose \textbf{D}omain \textbf{S}pecialization for \textbf{T}ask \textbf{O}riented \textbf{D}ialog (DS-TOD), a novel domain specialization framework for task-oriented dialog. 
DS-TOD, depicted in Figure~\ref{fig:DS-TOD}, has three steps: (1) we extract domain-specific terms (e.g., terms related to ordering \textit{taxi} or terms related to buying a \textit{train} ticket) from the training portions of a task-specific TOD corpus; (2) we next use the extracted terms to obtain domain-specific data from large unlabeled corpora (e.g., Reddit); (3) finally, we conduct intermediate training of a PLM (e.g., BERT) on the domain-specific data in order to inject the domain-specific knowledge into the encoder. As a result, we obtain a domain-specialized PLM, which can then be fine-tuned for downstream TOD tasks, e.g., dialog state tracking. 


\paragraph{Contributions.} We advance the state-of-the-art in TOD with the following contributions: \textbf{(i)} Departing from general-purpose dialogic pretraining~\citep[e.g.,][]{henderson-etal-2019-repository}, we leverage a simple terminology extraction method to construct \textsc{DomainCC} and \textsc{DomainReddit} corpora which we then use for domain-specific LM and dialogic pretraining, respectively. 
 \textbf{(ii)} We examine different objectives for injecting domain-specific knowledge into PLMs: we empirically compare Masked Language Modeling (MLM) applied on the ``flat'' domain dataset \textsc{DomainCC} against two different Response Selection (RS) objectives \cite{henderson-etal-2019-training, oord2018representation} applied on the dialogic \textsc{DomainReddit} corpus.
 We demonstrate the effectiveness of our specialization on two TOD tasks -- dialog state tracking (DST) and response retrieval (RR) -- for five domains from the \textsc{MultiWOZ} dataset~\cite{budzianowski-etal-2018-multiwoz,eric-etal-2020-multiwoz}. 
 \textbf{(iii)} We propose modular domain specialization for TOD via \textit{adapter} modules~\cite{pmlr-v97-houlsby19a,pfeiffer-etal-2020-mad}. Additional experiments reveal the advantages of adapter-based specialization in \textit{multi-domain} TOD: combining domain-specific adapters via stacking~\cite{pfeiffer-etal-2020-mad} or fusion \cite{pfeiffer-etal-2021-adapterfusion} (a) performs \emph{en par} with or outperforms expensive multi-domain pretraining, while (b) having a much smaller computational footprint.\footnote{Assume $N$ mutually close domains and a bi-domain downstream setup (any two domains). With an adapter-based approach, we pretrain one adapter for each domain (complexity: $N$) and then combine the adapters of the two domains intertwined in the concrete downstream setup. In contrast, multi-domain specialization would require one bi-domain pretraining for each two-domain combination (complexity: $N^2$).} 
 
\section{Data Collection}
\label{sec:data}

\begin{table*}[th]
\def\arraystretch{0.9}
\centering
\resizebox{\textwidth}{!}{%
\begin{tabular}{lccccc}
\toprule
 \textbf{} & \textbf{Taxi} & \textbf{Restaurant} & \textbf{Hotel} & \textbf{Train} & \textbf{Attraction}  \\ \midrule
Slot names &
  \begin{tabular}[c]{@{}c@{}} \emph{destination}, \emph{departure}, \\ \emph{arriveBy}, \emph{leaveAt}\end{tabular} &
  \begin{tabular}[c]{@{}c@{}}\emph{pricerange}, \emph{area}, \\ \emph{day}, \emph{people}, \emph{food}, \\ \emph{name}, \emph{time}\end{tabular} &
  \begin{tabular}[c]{@{}c@{}}\emph{pricerange}, \emph{parking}, \\ \emph{internet}, \emph{stars}, \emph{area}, \\ \emph{type}, \emph{people}, \emph{day}, \\ \emph{stay}, \emph{name}\end{tabular} &
  \begin{tabular}[c]{@{}c@{}}\emph{destination}, \emph{departure}, \\ \emph{day}, \emph{people}, \emph{arriveBy}, \\ \emph{leaveAt}\end{tabular} &
  \begin{tabular}[c]{@{}c@{}}\emph{area}, \emph{type}, \\ \emph{name}\end{tabular} \\
  \midrule
  \begin{tabular}[l]{@{}l@{}}\# Total (tr.,~dev,~test)\end{tabular}  & 1654, 207, 195 & 3813, 438, 437 & 3381, 416, 394   & 3103, 484, 494 & 2717, 401, 395 \\
\begin{tabular}[l]{@{}l@{}}\# Multi-domain (tr.,~dev,~test)\end{tabular}  & 1329, 150, 143 & 2616, 388, 375 & 2868, 360, 327   & 2828, 454, 461 & 2590, 390, 383 \\
\begin{tabular}[l]{@{}l@{}}\# Single domain (tr.,~dev,~test)\end{tabular} & 325, 57, 52    & 1197, 50, 62   & 513, 56, 67      & 275, 30, 33    & 127, 11, 12    \\
\cmidrule(){2-6}
\% Single domain                                                       & 24.62\%        & 19.00\%        & 15.21\% & 7.25\%         & 3.49\%    
\\
\bottomrule
\end{tabular}%
}
\caption{Statistics for MultiWOZ 2.1 dataset. For each domain, we report slot names, the total number of dialogs as well as the number of single-domain and multi-domain dialogs.}
\label{tab:multiwoz_domain}
\vspace{-0.5em}
\end{table*}


We create large-scale domain-specific corpora in two steps: given a collection of in-domain dialogs we first extract salient domain terms (\S\ref{domain-specific-term-extraction}); we then use these domain terms to filter content from CCNet~\cite{wenzek-etal-2020-ccnet} as a large general corpus and Reddit as a source of dialogic data (\S\ref{indomain-retrieval}). 
\subsection{Domain-Specific Ngrams}
\label{domain-specific-term-extraction}

\begin{table*}[t]
\def\arraystretch{0.92}
\centering
\resizebox{\textwidth}{!}{%
\begin{tabular}{ll}
\toprule
\textbf{Domain} & \textbf{Ngrams}\\
\midrule 
Taxi &
  \begin{tabular}[c]{@{}l@{}}
  \textit{taxi, contact number, book a taxi, booked, time schedule, pickup, leaving, booked type, booking completed, departing, destination,} \\ 
  \textit{cab, completed booked, honda, ford, audi, lexus, toyota, departure, skoda, lexus contact, toyota contact, ford contact, volvo, } \\              
  \textit{train station, departure site, tesla, audi contact, honda contact, skoda contact, picking, departing, volkswagen}\end{tabular} \\
 \midrule
Attraction &
  \begin{tabular}[c]{@{}l@{}}
  \textit{museum, college, entrance, attraction, information, centre town, center town, entertainment, swimming pool, gallery, sports, } \\              
  \textit{nightclub, pounds, park, postcode, architecture, centre area, center area, cinema, church, trinity college, entrance free,} \\              
  \textit{jello gallery, post code, town centre, town center, downing college} 
  \end{tabular} \\
  \midrule
Train &
  \begin{tabular}[c]{@{}l@{}}
  \textit{train station, travel time, leaving, pounds, train ticket, departing, payable, train leaving, cambridge, london, reference id, }\\              
 \textit{arrive, destination, kings cross, total fee, departure, arriving, book a train, booked, stansted, stansted airport, peterborough, }\\
  \textit{traveling, trip, airport, booking successful, norwich}\\
  \end{tabular} \\
 \midrule
Hotel &
  \begin{tabular}[c]{@{}l@{}}
  \textit{hotel, nights, parking, free parking, wifi, star hotel, price range, free wifi, guesthouse, guest house, internet, guest, hotel room, } \\
  \textit{star rating, expensive room, priced, rating, book room, moderately priced, moderate price, stay for, reservation, breakfast available, } \\
 \textit{book people, fully booked, booking,  reference}\end{tabular} \\
  \midrule
Restaurant &
  \begin{tabular}[c]{@{}l@{}}\textit{restaurant, food, price range, expensive, cheap, priced, chinese food, italian food, moderately priced, south town, book table, } \\
  \textit{city, north town, serving, city centre, city center, european food, reservation, food type, phone address, centre town, center town, } \\
  \textit{expensive restaurant, moderate price, cuisine, restaurant center, restaurant centre, south town, expensive price, east town, } \\
  \textit{cheap restaurant, indian food, asian food, british food, book people}\end{tabular}\\
\bottomrule
\end{tabular}%
}
\caption{\label{tab:domain_terms} Salient domain ngrams extracted from the single-domain training portions of MultiWOZ.}
\end{table*}


We start from Wizard-of-Oz, a widely used multi-domain TOD dataset ~\citep[MultiWOZ;][]{budzianowski-etal-2018-multiwoz}: we resort to the revised version 2.1~\citep{eric-etal-2020-multiwoz} and work with the five domains that have test dialogs: \textit{Taxi}, \textit{Attraction}, \textit{Train}, \textit{Hotel}, and \textit{Restaurant}. Table~\ref{tab:multiwoz_domain} shows the statistics of domain-specific MultiWOZ subsets.


To obtain large domain-specific corpora for our intermediate training, we first construct sets of domain-specific ngrams for each domain. To this end, we first compute TF-IDF scores for all \{1,2,3\}-grams found in single-domain dialogs from MultiWOZ training sets\footnote{E.g., for the \textit{Taxi} domain, we collect all training dialogs that span only that domain (i.e., only taxi ordering) and omit dialogs that besides \textit{Taxi} involve one or more other domains (e.g., taxi ordering and hotel booking in the same dialog).}: our term frequency (TF) is the total ngram frequency in all domain dialogs; the inverse document frequency (IDF) is here the inverse of the proportion of dialogs that contain the ngram. We then select $N$ ngrams with the largest TF-IDF scores (in all our experiments, we set $N = 80$) and manually eliminate from the list ngrams that are not intrinsic to the domain (e.g., weekdays, named locations). Finally, since MultiWOZ terms follow the British English spelling (e.g., \textit{centre}, \textit{theatre}), we add the corresponding American English word forms (e.g., \textit{center}, \textit{theater}). The complete resulting ngram sets for all domains are given in Table~\ref{tab:domain_terms}.


\subsection{Domain-Specific Corpora}%
\label{indomain-retrieval}
We next use the extracted domain ngrams to retrieve two types of in-domain data for domain specialization: (i) flat text and (ii) dialogic data.     


\paragraph{\textsc{DomainCC}.} 
For each of the five MultiWOZ domains, we create the corresponding flat text corpus for MLM training by filtering out 200K sentences from the English portion of {CCN}et~\citep{wenzek-etal-2020-ccnet} -- a high-quality collection of monolingual corpora extracted from 
CommonCrawl\footnote{\url{https://commoncrawl.org/}} 
that has been used for pretraining multilingual PLMs~\citep{conneau-etal-2020-unsupervised, liu-etal-2020-multilingual-denoising} -- that contain one or more of the previously extracted domain terms. We additionally clean all \textsc{DomainCC} portions by removing email addresses and URLs, and lower-casing all terms. We provide example excerpts for each domain in the Appendix.



\paragraph{\textsc{DomainReddit}.}
%
%
%
%
\begin{table}[t!]
\def\arraystretch{0.94}
\centering
\resizebox{0.48\textwidth}{!}{%
\begin{tabular}{lcl}
\toprule \textbf{Subreddit} & \textbf{\# Members} & \textbf{Domains} \\ \midrule
travel & 5.8M & Taxi, Attraction, Train, Hotel, Restaurant \\
backpacking & 2.5M & Taxi, Attraction, Train, Hotel, Restaurant \\
solotravel & 1.7M & Taxi, Attraction, Train, Hotel, Restaurant \\
CasualUK & 797K & Taxi, Attraction, Train, Hotel, Restaurant \\
unitedkingdom & 553K & Taxi, Attraction, Train, Hotel, Restaurant \\
restaurant & 81.6K & Restaurant \\
trains & 64.8K & Train, Attraction\\
hotel & 1.8K & Hotel\\
hotels & 4.9K & Hotel\\
tourism & 3.9K & Taxi, Attraction, Train, Hotel, Restaurant\\
uktravel & 1.5K & Taxi, Attraction, Train, Hotel, Restaurant\\
taxi & 0.6K & Taxi\\
\bottomrule
\end{tabular}%
}
\caption{\label{tab:subreddit} Subreddits and associated domains selected for creating \textsc{DomainReddit}.}
\vspace{-0.75em}
\end{table}
%
%
%
%
%
%
%
%
\begin{table*}[t]
\setlength{\tabcolsep}{5pt}
\small{
\begin{tabularx}{\linewidth}{lX}
\toprule
 \textbf{Field}                      & \textbf{Example}    \\\midrule
Subreddit              & restaurant \\\midrule
Context & \begin{tabular}[c]{@{}l@{}}
  \textit{Hosts don’t get tips? That’s news to me. Most host positions} \textit{in my area get at least 1\% of sales;}\\ \textit{they make anywhere} \textit{between $60-$100 per night in tips!} \end{tabular}\\
  \midrule
Response &
\begin{tabular}[c]{@{}l@{}}
  \textit{We get tips but definitely not that much (in my experience). The tip out in my restaurant is 1\% split} \\ \textit{between shift leaders, food runners, and any other FOH other than servers/bartenders. Full time hosts} \\ \textit{get about 50-75 every other week} \end{tabular}\\
  \midrule
False Response &
\begin{tabular}[c]{@{}l@{}}
  \textit{Wow that’s terrible. Then again, my restaurant is in CA, so wages and guest check averages are} \\ \textit{usually higher.} \end{tabular}\\
\bottomrule
\end{tabularx}%
}
\caption{Example from \textsc{DomainReddit} dataset.}
\label{tab:reddit_example}
\vspace{-0.75em}
\end{table*}
Being constructed from CommonCrawl, \textsc{DomainCC} portions do not exhibit any natural conversational structure, encoding of which has been shown beneficial for downstream TOD \cite{henderson-etal-2019-training,wu-etal-2020-tod}. We thus additionally create a dialogic corpus for each domain: we employ the Pushshift API~\citep{baumgartner2020pushshift} to extract dialogic data from Reddit (period 2015--2019). 
To this end, we select subreddits related to \textit{traveling} (listed in Table~\ref{tab:subreddit}) which we believe align well with the content of MultiWOZ, which was created by simulating conversations between tourists and clerks in a tourist information center. 
%
%
Each of the subreddits contains threads composed of a series of comments, each of which can serve as a \textit{context}  followed by a series of \textit{responses}. For \textsc{DomainReddit} we select context-response pairs where either the context utterance or the response contains at least one of the domain-specific terms. To construct examples for injecting conversational knowledge, we follow \citet{henderson-etal-2019-repository} and couple each \textit{true} context-response pair (i.e., a comment and its immediate response) with a \textit{false response} -- a non-immediate response from the same thread. 
Table~\ref{tab:reddit_example} provides an example context with its true and one false response; further examples, for all domains, are available in the Appendix. 
Finally, we also clean \textsc{DomainReddit} by removing email addresses and URLs as well as comments having fewer than 10 characters. The total number of Reddit triples \textit{(context, true response, false response)} that we extract this way for the MultiWOZ domains is as follows: \textit{Taxi} -- 120K; \textit{Attraction} -- 157K; \textit{Hotel} -- 229K; \textit{Train} -- 229K; and \textit{Restaurant} -- 243K.



\section{Domain Specialization Methods}
\label{sec:method}
The next step in DS-TOD is the injection of domain-specific knowledge through intermediate model training on \textsc{DomainCC} and \textsc{DomainReddit}. To this end, we train a PLM (1) via Masked Language Modeling on \textsc{DomainCC} and (2) using two different Response Selection objectives on \textsc{DomainReddit}. Finally, for all objectives, we compare full domain fine-tuning (i.e., we update all PLM parameters) against adapter-based specialization where we freeze the PLM parameters and inject domain knowledge into new adapter layers.

\subsection{Training Objectives}
\label{ss:objectives}

\paragraph{Masked Language Modeling (MLM).}
Following successful work on domain-adaptive pretraining via LM~\citep{gururangan-etal-2020-dont, aharoni-goldberg-2020-unsupervised,glavas-etal-2020-xhate}, we investigate the effect of running standard MLM on the domain-specific portions of \textsc{DomainCC}.


\paragraph{Response Selection (RS).}
RS objectives force the model to recognize the correct response utterance given the context -- pretraining with such objectives is particularly useful for conversational settings, including TOD tasks~\citep{henderson-etal-2019-training, henderson-etal-2020-convert}.
We consider two RS objectives. The first is a simple pairwise binary classification formulation (\textbf{RS-Class}): given a context-response pair, predict whether the response is a true (i.e., immediate) response to the context. We straightforwardly use pairs of contexts and their true responses from \textsc{DomainReddit} as positive training instances. Next, we create negative samples for each positive instance as follows: (a) we use the crawled \textit{false response} from \textsc{DomainReddit}, which represents a relevant but non-consecutive response from the same thread; such non-immediate responses from the same thread represent the so-called \textit{hard negatives} introduced to prevent the model from learning simple lexical cues and similar heuristics that poorly generalize; (b) we additionally randomly sample $k$ utterances from the same domain but different threads (these represent the so-called \textit{easy negatives}).\footnote{$k$ is uniformly sampled from the set \{1, 2, 3\}.} 


The second response selection objective (\textbf{RS-Contrast}) that we adopt is a type of loss function used for contrastive model training based on the representational similarities between sampled positive and negative pairs \cite{oord2018representation}. It has been used for pretraining cross-lingual language models~\citep{chi-etal-2021-infoxlm} and shown to be useful in information retrieval~\citep{reimers-gurevych-2021-curse,thakur2021beir,litschko2022cross}. The goal is to estimate the mutual information between pairs of variables by discriminating between a positive pair and its associated $N$ negative pairs. Given a true context-response pair and $N$ negatives (same as for RS-Class), the noise-contrastive estimation (NCE) loss is computed as:  
%
%
\begin{equation*}
L_{\mathit{NCE}} = -\log\frac{\exp\left(f(c, r_+)\right)}{\sum_{i = 1}^{N+1}\exp\left(f(c, r_i)\right)}\,,
\end{equation*}

\normalsize \noindent where $c$ is the context, $r_+$ is the true response and $r_i$ iterates over all responses for the context -- the true response $r_+$ and $N$ false responses; a function $f$ produces a score meant to indicate whether the response $r$ is a true response of the context $c$. 

By learning to differentiate whether the response is true or false for a given context (RS-Class) or to produce a higher score for a true response than for false responses (RS-Contrast), RS objectives encourage the PLM to adapt to the underlying structure of the conversation. By feeding only in-domain data to it, we impart domain-specific conversational knowledge into the model.


\subsection{Adapter-Based Domain Specialization}
\label{ss:adapters}

Fully fine-tuning the model requires adjusting all of the model's parameters, which can be undesirable due to large computational effort and risk of catastrophic forgetting of the previously acquired knowledge~\citep{mccloskey1989catastrophic, pfeiffer-etal-2021-adapterfusion}. To alleviate these issues, we investigate the use of adapters~\citep{pmlr-v97-houlsby19a}, additional parameter-light modules that are injected into a PLM before fine-tuning. In adapter-based fine-tuning only adapter parameters are updated while the pretrained parameters are kept frozen (and previously acquired knowledge thus preserved). We adopt the adapter-transformer architecture proposed by \citet{pfeiffer-etal-2020-mad}, which inserts a single adapter layer into each transformer layer and computes the output of the adapter, a two-layer feed-forward network, as follows:  
%
%
%
\begin{equation*}
    \textit{Adapter}(\bm{h}, \bm{r}) = U \cdot g(D \cdot \bm{h}) + \bm{r},
\end{equation*}
\noindent with $\bm{h}$ and $\bm{r}$ as the hidden state and residual of the respective transformer layer. $D \in \mathbb{R}^{m \times h}$ and $U \in \mathbb{R}^{h \times m}$ are the linear down- and up-projections, respectively ($h$ being the transformer's hidden size, and $m$ as the adapter's bottleneck dimension), and $g(\cdot)$ is a non-linear activation function. The residual $\bm{r}$ is the output of the transformer's feed-forward layer whereas $\bm{h}$ is the output of the subsequent layer normalization. The down-projection $D$ compresses token representations to the adapter size $m \ll h$, and the up-projection $U$ projects the activated down-projections back to the transformer’s hidden size $h$. The ratio $h/m$ captures the factor by which the adapter-based fine-tuning is more parameter-efficient than full fine-tuning.

For multi-domain TOD scenarios (i.e., dialogs covering more than a single domain), we further experiment with combinations of individual domain adapters: (1) sequential stacking of adapters one on top of the other \citep{pfeiffer-etal-2020-mad} and (2) adapter fusion, where we compute a weighted average of outputs of individual adapter, with fusion weights as parameters that are learned in the final task-specific fine-tuning~\citep{pfeiffer-etal-2021-adapterfusion}. 



\section{Experiments}
\setlength{\tabcolsep}{4pt}
\label{sec:experiments}
\begin{table*}[t]
\centering
\scriptsize{
\begin{tabular}{l ccccc g ccccc g}
\toprule
& \multicolumn{6}{c}{\textbf{Dialog State Tracking}}     & \multicolumn{6}{c}{\textbf{Response Retrieval}}          \\ 
\textbf{Model} & \textbf{Taxi} & \textbf{Restaurant} & \textbf{Hotel} & \textbf{Train} & \textbf{Attraction} & \textbf{Avg.} &  \textbf{Taxi} & \textbf{Restaurant} & \textbf{Hotel} & \textbf{Train} & \textbf{Attraction} & \textbf{Avg.}\\ \cmidrule(lr){2-7}\cmidrule(lr){8-13}
BERT                 & 23.87 & 35.44      & 30.18 & 41.93 & 29.77      & 32.24 & 23.25 & 37.61      & 38.97 & 44.53 & 48.47      & 38.57 \\
TOD-BERT             & 30.45 & 43.58      & 36.20 & 48.79 & 42.70      & 40.34 & 45.68 & 57.43      & 53.84 & \textbf{60.66} & 60.26      & 55.57 \\\midrule
BERT-MLM             & 23.74 & 37.09      & 32.77 & 40.96 & 36.66      & 34.24 &   31.37    &        53.08    &     45.41  &   51.66    &       52.23     &     46.75   \\
TOD-BERT-MLM         & 29.94 & 43.14      & 36.11 & 47.61 & 41.54      & 39.67 & 41.77 & 55.27      & 50.60  & 55.17 & 54.62      & 51.49 \\
TOD-BERT-RS-Class          & \textbf{36.39} & 43.38      & 37.89 & 48.82 & 43.31      & 41.96 & 47.01 & 58.21      & \textbf{57.05} & 59.70 & 57.72      & 55.94 \\
TOD-BERT-RS-Contrast     & 35.03 & \textbf{44.81}      & \textbf{38.74} & \textbf{49.04} & 42.73      & \textbf{42.07} & 48.04 & 59.82      & 54.49 & 60.06 & 60.63      & 56.61 \\\midrule
BERT-MLM-adapter     & 22.52 & 40.49      & 31.90 & 42.17 & 35.05      & 34.43 &  32.84     & 44.01           &    39.15   &  38.43     &  45.05          &   39.90     \\
TOD-BERT-MLM-adapter & 32.06 & 44.06      & 36.74 & 48.84 & \textbf{43.50}      & 41.04 & 49.08 & 58.18      & 55.55 & 59.46 & 60.26      & 56.51 \\
TOD-BERT-RS-Class-adapter  & 33.10 & 42.57      & 38.61 & 49.03 & 42.35      & 41.13 & \textbf{49.59} & \textbf{61.26}      & 56.87 & 58.88 & 60.00      & \textbf{57.32} \\
TOD-BERT-RS-Contrast-adapter &
  34.90 &
  44.42 &
  37.52 &
  48.71 &
  42.83 &
  41.68 &
  47.97 &
  58.97 &
  55.41 &
  59.15 &
  \textbf{61.95} &
  56.69 \\
  \bottomrule
\end{tabular}%
}
\caption{Results of DS-TOD models on two downstream tasks: Dialog State Tracking (DST) and Response Retrieval (RR) with joint goal accuracy (\%) as the metric for DST and $\textsc{R}_{100}@1$ \cite{henderson-etal-2020-convert} (\%) for RR.}
\label{tab:eval_result}
\vspace{-0.75em}
\end{table*}

We demonstrate the effectiveness of our domain-specialization framework by comparing it to non-specialized baseline models and thoroughly compare different specialization methods from \S\ref{sec:method}.

\paragraph{Evaluation Task and Measures.} 
We evaluate our domain-specialized models and baselines on two prominent downstream TOD tasks: \emph{dialog state tracking (DST)} and \emph{response retrieval (RR)}. DST is treated as a multi-class classification task based on a predefined ontology, where given the dialog history, the goal is to predict the output state, i.e., (domain, slot, value) tuples. For our implementation, we follow \citet{wu-etal-2020-tod}, and represent the dialog history as a sequence of utterances. The model then needs to predict slot values for each (domain, slot) pair at each dialog turn. We report the \textit{joint goal accuracy}, in which the predicted dialog states are compared to the ground truth slot values at each dialog turn. The ground truth contains slot values for all the (domain, slot) candidate pairs. A prediction is considered correct if and only if all predicted slot values exactly match its ground truth values. RR is a ranking problem, relevant for retrieval-based  TOD systems~\citep{wu-etal-2017-sequential, henderson-etal-2019-training}. Following \newcite{henderson-etal-2020-convert} and \newcite{wu-etal-2020-tod}, we adopt recall at top rank given 100 randomly sampled candidates ($\textsc{R}_{100}@1$) as the evaluation metric for RR. 


\paragraph{Data.} In the pretraining procedure, we use the domain-specific portions of our novel \textsc{DomainCC} and \textsc{DomainReddit} resources (\S\ref{sec:data}). For the MLM training, we randomly sample 200K domain-specific contexts from \textsc{DomainCC} and dynamically mask 15\% of the subword tokens. For RS-Class and RS-Contrast, we randomly sample 200K instances from \textsc{DomainReddit}. We evaluate the efficacy of the methods on DST and RR using MultiWOZ 2.1~\citep{eric-etal-2020-multiwoz}. Since we aim to understand the effect of the domain specialization, we construct domain-specific training, development, and testing portions from the original data set by assigning them all dialogs that belong to a domain (i.e., both single- and multi-domain dialogs) from respective overall (train, dev, test) portions.

\paragraph{Models and Baselines.} We experiment with two PLMs: BERT~\citep{devlin-etal-2019-bert} and its TOD-sibling, TOD-BERT~\citep{wu-etal-2020-tod}.\footnote{We use the pretrained models \texttt{bert-base-cased} and \texttt{TODBERT/TOD-BERT-JNT-V1} from HuggingFace.}
As baselines, we report the performance of the non-specialized variants and compare them against our domain-specialized PLM variants, obtained after intermediate MLM-training on \textsc{DomainCC} or RS-Class/RS-Contrast training on \textsc{DomainReddit}.


\paragraph{Hyperparameters and Optimization.} During domain-specific pretraining, we fix the maximum sequence length to $256$ subword tokens (for RS objectives, we limit both the context and response to $128$ tokens). We train for $30$ epochs, in batches of $32$ instances and search for the optimal learning rate among the following values: $\{1\cdot 10^{-4}, 5\cdot 10^{-5}, 1\cdot 10^{-5}, 1\cdot 10^{-6}\}$. We apply early stopping based on development set performance (patience:~3~epochs). We minimize the cross-entropy loss using Adam~\citep{kingma2014adam}. 
%
%
For downstream evaluation, we train for 300 epochs in batches of $6$ (DST) and $24$ instances (RS) with the learning rate fixed to $5\cdot 10^{-5}$. We also apply dev-set-based early stopping (patience: 10 epochs).

\section{Results and Discussion}
\label{sec:discussion}


\paragraph{Overall performance.}

We report downstream DST and RR results in Table~\ref{tab:eval_result}, which is segmented in three parts: (1) at the top we show the baseline results (BERT, TOD-BERT) without any domain specialization; (2) in the middle of the table we show results of PLMs domain-specialized via full fine-tuning; (3) the bottom of the table contains results for our adapter-based domain specialization.

In both DST and RR, TOD-BERT massively outperforms BERT due to its conversational knowledge.
Domain specialization brings gains for both PLMs across the board. The only exception is full MLM-fine-tuning of TOD-BERT (i.e., TOD-BERT-MLM vs. TOD-BERT; -4\% for RR and -0.8\% for DST): we believe that this is an example of negative interference -- while TOD-BERT is learning domain knowledge, it is -- because of MLM-based domain training -- forgetting the conversational knowledge obtained in dialogic pretraining \cite{wu-etal-2020-tod}. This hypothesis is further supported by the fact that adapter-based MLM specialization of TOD-BERT -- which prevents negative interference by design -- brings slight performance gains (i.e., TOD-BERT-MLM-adapter vs. TOD-BERT; +0.8\% for DST and +1.0\% for RR) and is consistent with the concurrent findings of \citet{qiu2021different}.    

Overall, domain specialization with RS seems to be more robust than that via MLM-ing, with the two variants (RS-Class and RS-Contrast) exhibiting similar average performance across evaluation settings. This points to the importance of injecting both the knowledge of dialogic structure as well as domain knowledge for performance gains in TOD tasks in the domain of interest. 



Interestingly, the gains from domain specialization are significantly more pronounced for \textit{Taxi} than for other domains. We relate this to the proportion of the single-domain dialogs for a given domain in MultiWOZ, which is by far the largest (24\%, see Table~\ref{tab:multiwoz_domain}) for the \textit{Taxi} domain. Consequently, successful specialization for that domain is \textit{a priori} more likely to show substantial gains on MultiWOZ (i.e., less multi-domain influence). 

An encouraging finding is that, on average, adapter-based specialization yields similar gains as specialization via full fine-tuning: given that adapter fine-tuning is substantially more efficient, this holds the promise of more sustainable TOD.

\begin{figure}[t]
	\centering
  \includegraphics[width=0.48\textwidth]{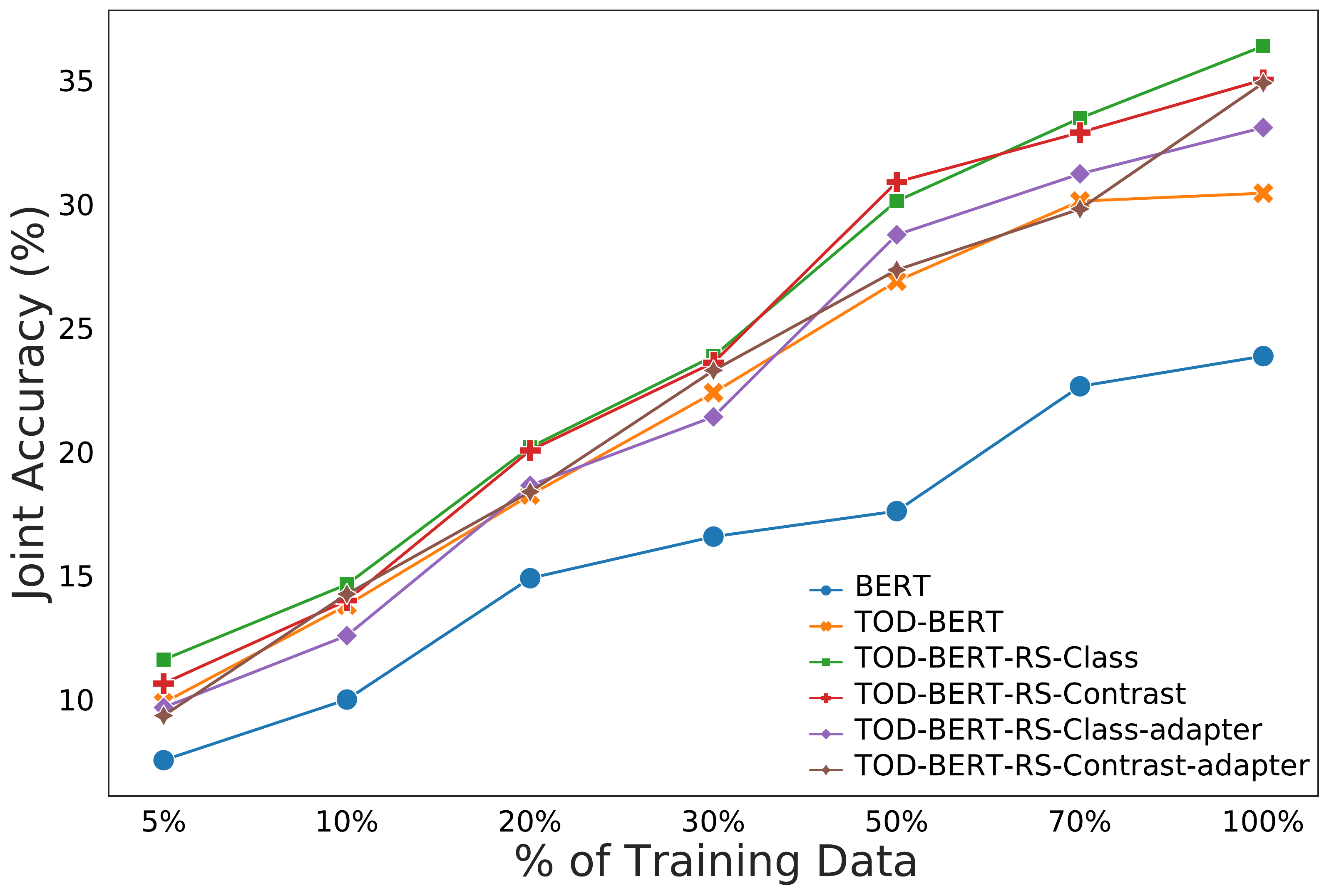}
	\caption{Sample efficiency of DS-TOD for DST: joint goal accuracy (\%) for randomly sampled sub-portions (5\%, 10\%, 20\%, 30\%, 50\%, 70\%, and 100\%) of the downstream training data from the \textit{Taxi} domain.}
	\label{fig:few_shot_taxi}
	\vspace{-0.5em}
\end{figure}

\paragraph{Sample Efficiency.} To further understand the effect of the injected domain-specific knowledge, we conduct an additional few-shot analysis (Figure~\ref{fig:few_shot_taxi}) on DST. To this end, we select the \textit{Taxi} domain, since we witnessed the largest gains for that domain. We analyse the differences in performance between baseline and domain-specialized PLMs when they are exposed to downstream training portions of different sizes, ranging from 5 to 100\% of the whole training dataset.\footnote{Note that 5\% of the training data in the \textit{Taxi} domain amounts to 83 dialogs.} 
%
%
TOD-BERT retains a sizable performance gap over BERT for all settings, pointing to the power of dialogic pretraining. Importantly, for all dataset sizes, the performances of the domain-specialized variants of TOD-BERT-RS-\{Class, Contrast\} surpass the one of the non-specialized TOD-BERT. Even more interestingly, specialized variants exposed to only 50\% of the DST training data manage to surpass the performance of TOD-BERT fine-tuned on all of the training data (100\%). This suggests that self-supervised domain specialization has the potential to substantially reduce the amount of annotated TOD data required to reach some performance level.

\begin{figure}[t]
	\centering
  \includegraphics[width=0.48\textwidth]{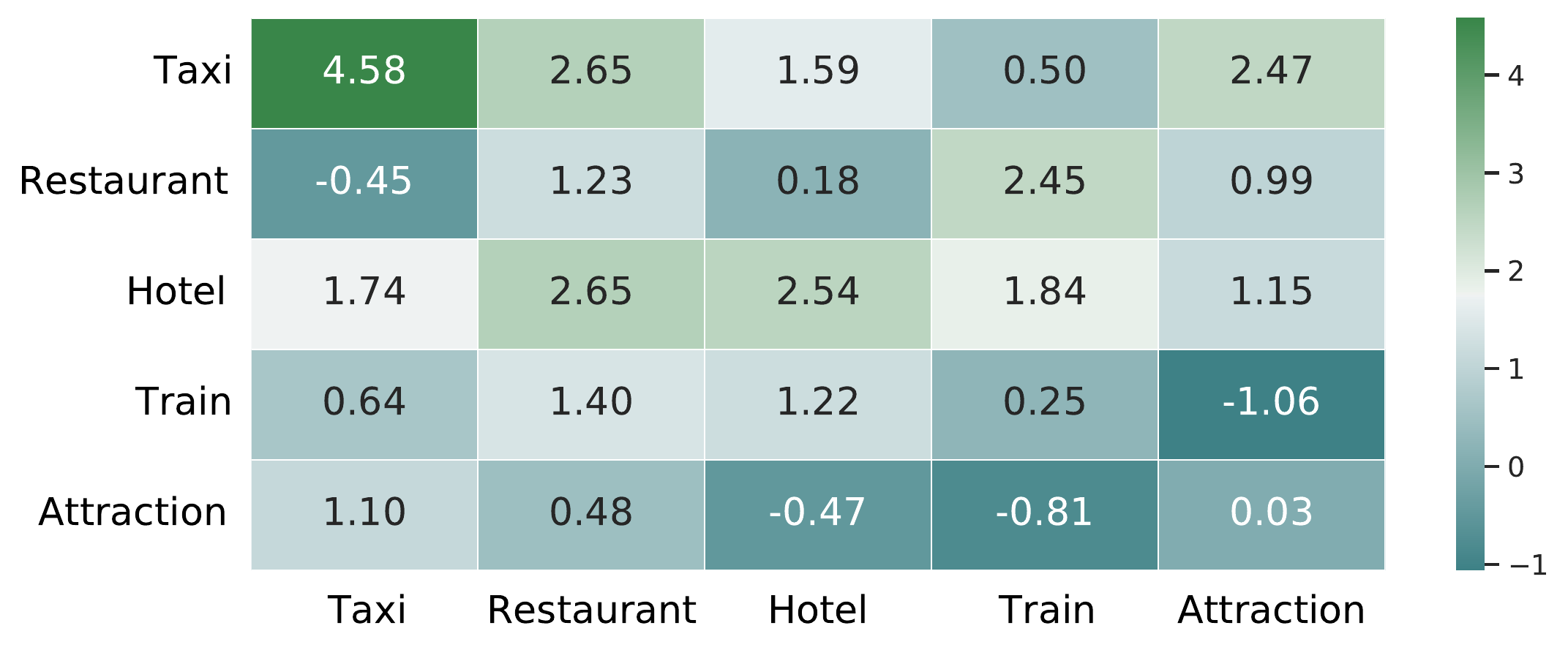}
	\caption{Relative improvements (TOD-BERT-RS-Contrast~vs.~TOD-BERT) in cross-domain DST transfer.}
	\label{fig:domain_transfer_diff}
	\vspace{-0.5em}
\end{figure}

\paragraph{Cross-Domain Transfer.} MultiWOZ domains are mutually quite related: some are similar, i.e., share vocabulary and slots (e.g., \textit{Taxi} and \textit{Train}) whereas others often appear together in a dialog (e.g., \textit{Train} and \textit{Hotel}; see Table \ref{tab:multiwoz_domain} for the number of multi-domain MultiWOZ dialogs).
We thus next investigate whether intermediate training for one domain benefits other, closely related domains. To this end, we expose models specialized for one domain (e.g., \textit{Taxi}) to downstream fine-tuning and evaluation in the other domain (e.g., \textit{Restaurant}). Figure~\ref{fig:domain_transfer_diff} summarizes the deltas in performance between the non-specialized TOD-BERT and TOD-BERT-RS-Contrast for all domain pairs. Encouragingly, the specialization for one domain seems to generally lead to downstream gains in related domains too: the gains are most prominent for pairs of domains that frequently co-occur in dialogs -- \textit{Hotel} pretraining for the \textit{Restaurant} downstream (and vice versa) and \textit{Taxi} pretraining for downstream tasks in the \textit{Restaurant} and \textit{Attraction} domains.    

\begin{figure}[t]
	\centering
  \includegraphics[trim={0cm 0 0 0.1cm},clip, width=0.48\textwidth]{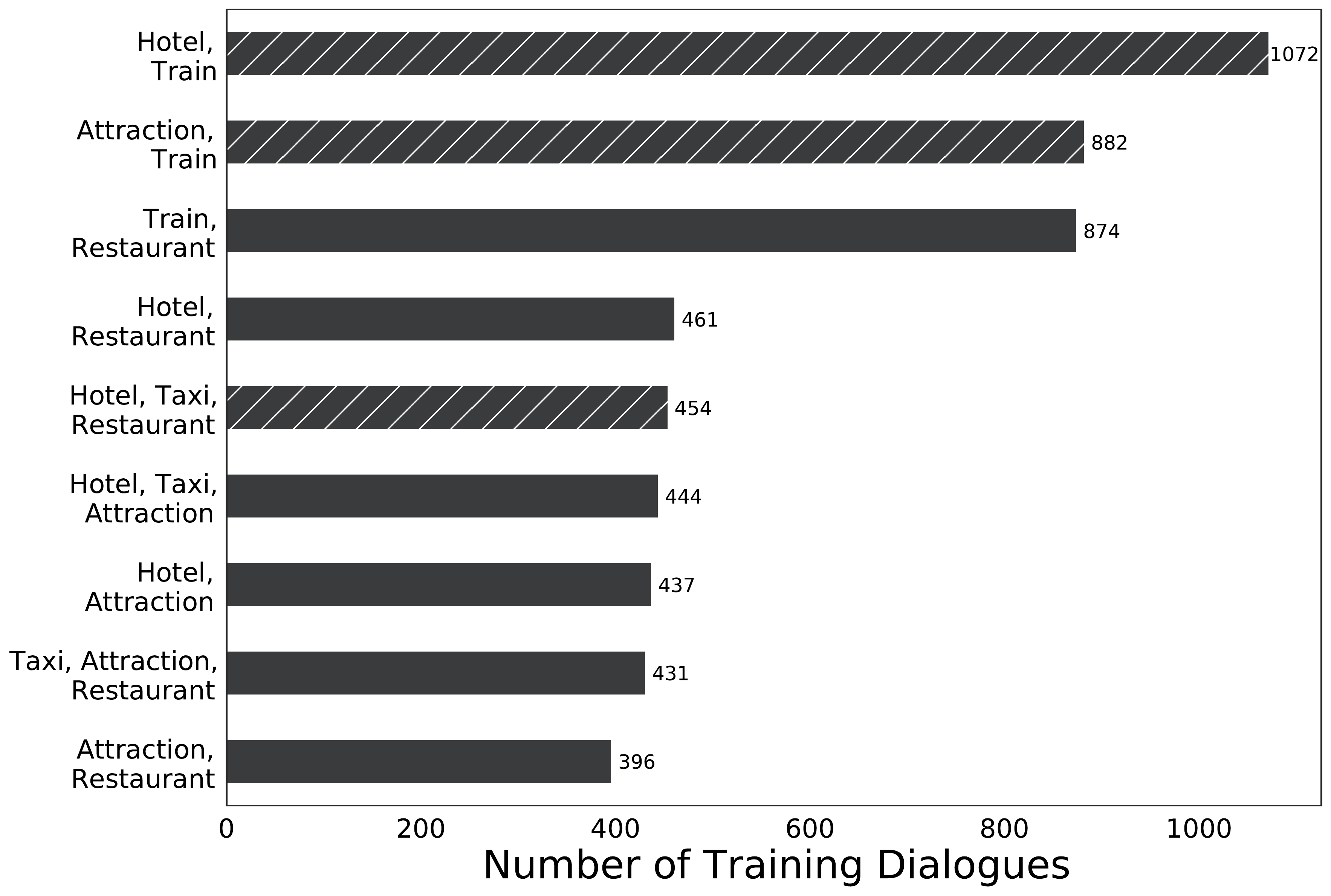}
	\caption{Number of dialogs in the MultiWOZ 2.1 training dataset when joining several domains. Striped bars indicate the domain combinations used for multi-domain specialization.}
	\label{fig:domain_dials}
	\vspace{-0.5em}
\end{figure}

\paragraph{Multi-Domain Specialization.} In many real-world scenarios, a single model needs to be able to handle multiple domains because (a) multi-domain (MD) dialogs exist and (b) simultaneous deployment of multiple single-domain (SD) models may not be feasible. To simulate this scenario, we conduct an additional analysis, in which we concatenate dialogs from respective MultiWOZ portions that cover concrete combinations of two or three domains. We choose three domain combinations with the largest number of MD dialogs, namely the two largest 2-domain combinations and the largest 3-domain combination (Figure~\ref{fig:domain_dials}): 
\textit{Hotel}+\textit{Train}, \textit{Attraction}+\textit{Train}, and \textit{Hotel}+\textit{Taxi}+\textit{Restaurant}.

As baselines, we report the performance of BERT and TOD-BERT fine-tuned on the respective MD TOD training sets. We test the effect of MD specialization in two variants: (1) \textit{fully specialized model trained for multiple domains} (\textbf{Full-FT}): as RS-Class has proven to be effective in our SD-specialization experiments, we run RS-Class training on the concatenation of the selected domains from \textsc{DomainReddit} that correspond to the domains of the joint training sets. Accordingly, the training data is roughly twice (or three times) as big as that used for SD specialization; (2) \textit{composition of SD adapters for multiple domains}: while for Full-FT, a new intermediate training is necessary for each domain combination, with adapter-based specialization we can simply combine the adapters of relevant domains in downstream fine-tuning. In this setup, we combine the SD adapters by sequentially stacking them \citep{pfeiffer-etal-2020-mad} (\textbf{Stacking}) or by fusing them, i.e., interpolating between their outputs \citep{pfeiffer-etal-2021-adapterfusion} (\textbf{Fusion}).


The MD specialization results are shown in Table~\ref{tab:adapterstack_fusion}. Interestingly, combining SD adapters in downstream training (via Stacking or Fusion) performs \emph{en par} with full-sized two-domain specialization on \textsc{DomainReddit} by means of RS-Class training. 
In contrast to TOD-BERT-RS-Class (Full-FT), which requires full retraining of the model on the unlabelled domain-specific corpora for each combination of the domains, combining SD adapters is much more efficient as it does not require any further intermediate domain training for domain combinations. In the 3-domain setup (\textit{Hotel+Taxi+Restaurant}), the Fusion approach even outperforms the full 3-domain specialization (TOD-BERT-RS-Class Full-FT) by 2 points.

Overall, we find that the adapter compositions provide a simple and effective way to combine information from several domain-specialized adapters, removing the need for additional MD specialization in the face of MD dialogs downstream. 
%

\setlength{\tabcolsep}{3.5pt}
\begin{table}[t]
\centering
\scriptsize{
\begin{tabular}{llccc}
\toprule
 \textbf{Model} &  \begin{tabular}[c]{@{}c@{}}\textbf{Special.}\\ \textbf{Method}\end{tabular} & \begin{tabular}[c]{@{}c@{}}\textbf{Hotel+}\\ \textbf{Train}\end{tabular}  & \begin{tabular}[c]{@{}c@{}}\textbf{Attraction+}\\ \textbf{Train}\end{tabular} &
 \begin{tabular}[c]{@{}c@{}}\textbf{Hotel+Taxi+}\\ \textbf{Restaurant}\end{tabular}   \\ \midrule
BERT & -- & 42.66    & 45.06    & 37.00    \\
TOD-BERT & -- & 46.38    & 46.40    & 42.47    \\
\midrule
\multirow{3}{*}{TOD-BERT-RS-Class } & Full-FT & \textbf{47.39}    & \textbf{47.33}    & 42.39   \\
\cmidrule{2-5}
& Stacking  & 47.19    & 46.68    & 42.15    \\
& Fusion & 44.25    & 45.57    & \textbf{44.02}                \\
\bottomrule
\end{tabular}%
}
\caption{DS-TOD performance on DST in multi-domain scenarios. We compare the fully multi-domain-specialized variant (Full-FT) of the TOD-BERT-RS-Class model with its variant that combines readily available single-domain adapters (Stacking and Fusion) on three multi-domain evaluation sets.}
\label{tab:adapterstack_fusion}
\vspace{-0.5em}
\end{table}

\section{Related Work}
\label{sec:rw}
\paragraph{TOD Datasets.}
Datasets for task-oriented dialog can be divided into single-domain~\citep{wen-etal-2017-network, mrksic-etal-2017-neural} and multi-domain ones~\citep{budzianowski-etal-2018-multiwoz, rastogi2019towards}. The latter are generally seen as closer to real-world situations and intended usages of personal assistants, where strict adherence to a single domain is unlikely. While downstream TOD datasets exist for specific domains, corresponding large(er)-scale datasets that would enable domain-specific pretraining have been limited to the general domain \cite{henderson-etal-2019-repository}. 

\paragraph{Pretrained Language Models in Dialog.}
The advantages of large-scale pretraining of deep language models on massive amounts of text~\citep{devlin-etal-2019-bert, radford2019language, lewis-etal-2020-bart}, ubiquitous in natural language tasks, have also spilled over to task-oriented dialog. Recent research focused on either (1) leveraging general-domain dialogic resources (e.g., Reddit, Twitter) in order to improve downstream TOD tasks~\citep{henderson-etal-2019-training, henderson-etal-2020-convert, zhang-etal-2020-dialogpt, bao-etal-2020-plato, liu2021pretraining} or (2) using TOD datasets to inject dialogic structure into PLMs~\citep{wu-etal-2020-tod, peng2021soloist, su2021multi}. Neither of the two, however, considers domain adaptation or domain-specific pretraining. 

\paragraph{Domain Adaptation and Knowledge Reuse.}
Common unsupervised approaches for extracting domain-specific portions of large general domain corpora, rely on term and document frequencies~\citep{kim-etal-2009-extracting}, learn a candidate retrieval-based classifier~\citep{glavas-etal-2020-xhate} or perform unsupervised domain clustering with PLMs~\citep{aharoni-goldberg-2020-unsupervised}. 
In this work, we address the lack of in-domain resources by creating large-scale domain-specific corpora -- flat as well as dialogic -- for the five domains of the MultiWOZ dataset using a simple TF-IDF based term filtering approach.

Intermediate training is the prevalent approach for injecting domain knowledge into PLMs, either as a step before the downstream task-specific fine-tuning \cite{glavas-etal-2020-xhate} or in parallel with it (i.e., in a multi-task training setup)~\citep{gururangan-etal-2020-dont}. 
In the narrower context of TOD, \citet{whang2020domain} present the lone effort on domain specialization for TOD: they focus on easier, single-domain TOD and investigate the specialization effect with a single task, response retrieval. In this work, in contrast, we focus on dialogic domain-specific pretraining and show its effectiveness in multi-domain TOD.  
For efficiency and to avoid catastrophic forgetting, adapter modules have been widely used for parameter-efficient fine-tuning of PLMs for new tasks~\citep{pmlr-v97-houlsby19a} and languages \cite{pfeiffer-etal-2020-mad}.
Non-destructive adapter compositions (e.g., stacking or fusion) can be beneficial if multiple knowledge facets, stored in separate adapters, need to be leveraged \citep{pfeiffer-etal-2020-mad, pfeiffer-etal-2021-adapterfusion}.

\section{Reproducibility}
\label{sec:down}
We provide the complete lists of keywords used for domain-specific corpus filtering in Table~\ref{tab:domain_terms}, and a transparent description of the filtering approach in \S\ref{domain-specific-term-extraction}. We also release the filtered corpora for each domain. This makes our approach completely transparent and fully reproducible. 
All resources developed as part of this work are publicly available at: \url{https://github.com/umanlp/DS-TOD}.
\vspace{-0.3em}

\section{Conclusion}
\label{sec:conc}
We introduced \textbf{DS-TOD} -- a novel framework for domain specialization of PLMs for task-oriented dialog. 
Given a collection of in-domain dialogs, we extract domain terms and use them to filter in-domain dialogic corpora. Our experimental study, on five domains of the MultiWOZ dataset, shows that domain specialization, especially by means of response selection objectives on the dialogic in-domain corpora, leads to consistent gains in TOD tasks: dialogue state tracking and response retrieval. 
We hope that our domain-specific resources catalyze research on domain specialization for TOD, especially for multi-domain setups. Our future efforts will focus on the joint domain- and language-specialization for task-oriented dialog.  
\vspace{-0.5em}


\section*{Acknowledgements}
\label{sec:acknow}
The work of Goran Glava\v{s} has been supported by the Multi2ConvAI project of MWK Baden-Württemberg. Simone Paolo Ponzetto has been supported by the JOIN-T 2 project of the Deutsche Forschungsgemeinschaft (DFG). Chia-Chien Hung has been supported by Multi2ConvAI (MWK BW) and JOIN-T 2 (DFG). The work of Anne Lauscher was funded by Multi2ConvAI and by the European Research Council (grant agreement No. 949944, INTEGRATOR).

\section*{Ethical Considerations}
In this work, we have focused on domain specialization of task-oriented dialog models. While the limited scope of this manuscript prevents us from an in-depth discussion of potential ethical issues related to conversational AI in general, we would like to highlight the sensitivity of many of the envisioned future applications and the corresponding potential harms, e.g., unfair stereotypical biases encoded in pretrained language models, both general purpose \cite{nadeem2021stereoset,lauscher2021sustainable} and conversational PLMs \cite{barikeri-etal-2021-redditbias} alike and exclusion and misrepresentation within the large spectrum of users' (gender) identities~\citep{dev-etal-2021-harms, lauscher2022welcome}, which might arise depending on the sociotechnical deployment environment. Further, in light of potential environmental harms arising from training large PLMs~\citep{strubell-etal-2019-energy}, we would like to point the reader to the computational efficiency of our adapter-based specialization approaches, specifically in a multi-domain setup.

\bibliography{references}
\bibliographystyle{acl_natbib}

\appendix

\clearpage
\onecolumn
\label{sec:appendix}
\section{Domain-Specific Corpora}
In this section, we show examples from both \textsc{DomainCC} (Table~\ref{tab:domaincc}) and \textsc{DomainReddit} (Table~\ref{tab:domainreddit}). Both these resources were created starting from the salient domain terms listed in Table~\ref{tab:domain_terms}.
\begin{table}[h]
\def\arraystretch{0.90}
\centering
\resizebox{\textwidth}{!}{%
\begin{tabular}{ll}
\toprule \textbf{Domain} & \textbf{``Flat'' Text} \\ \midrule
Taxi &   \begin{tabular}[c]{@{}l@{}}
   \textit{\underline{\textbf{Taxis}}: licensed black \underline{\textbf{cabs}} operate a 24-hour, 365 day service from directly outside the arrivals area of the terminal building. Each} \\
  \textit{\underline{\textbf{taxi}} can carry up to five passengers (some can carry up to eight), with luggage and all are able to take wheelchair passengers.}\\  \end{tabular}\\
  \midrule
Restaurant &   \begin{tabular}[c]{@{}l@{}}\textit{\underline{\textbf{Asian food}} is very easy to like because it hits your mouth very differently than \underline{\textbf{European food}} does. In \underline{\textbf{European food}}, there may} \\\textit{be two things to hit - maybe sweet and salty, maybe salty-savory, but Asian kind of works around, plus you have that distinct in the} \\\textit{evening, a five course wine tasting dinner will be served in a gastronomic 2 Michelin starred \underline{\textbf{restaurant}}.}\\
\end{tabular}\\
\midrule
Train &  \begin{tabular}[c]{@{}l@{}}
\textit{Getting to centre \underline{\textbf{London}} is very easy as it take only one underground \underline{\textbf{train}} and it takes only 20-25 minutes to get to Oxford Circus.}\\ \textit{\underline{\textbf{Stansted airport}} is only 31 minutes away and all major motorways (M1, M11, North circular) is 5-10 minutes away.}\\

\end{tabular}\\
\midrule
Hotel &   \begin{tabular}[c]{@{}l@{}} \textit{Beautifully restored 1920's \underline{\textbf{guesthouse}}, comfortable and spacious bedrooms, lush gardens to explore, friendly and super helpful host,}\\ \textit{secure \underline{\textbf{parking}}. What more could you ask for! I would definitely recommend 6 on Kloof.}\\
\end{tabular}\\
\midrule
Attraction &   \begin{tabular}[c]{@{}l@{}}
\textit{On 31 august we travelled to Ely by train from kings cross and visited the Cathedral's interesting stained glass \underline{\textbf{museum}}. We also}\\ \textit{visited Oliver Cromwell's house nearby and sat outside for lunch, an extra bonus as it was a beautiful summer day. There was also}\\ \textit{time to look around Ely's \underline{\textbf{town centre}} before heading home.}\\
\end{tabular}\\
\bottomrule
\end{tabular}%
}
\caption{\label{tab:domaincc} Example from \textsc{DomainCC} dataset, where the salient domain terms are marked as \underline{\textbf{bold}}. The texts are displayed in the original version, without correcting typos.}

\end{table}

\begin{table}[h]
\def\arraystretch{0.9}
\centering
\resizebox{\textwidth}{!}{%
\begin{tabular}{lll}
\toprule \textbf{Domain} & \textbf{Dialogic Data} & \\ \midrule 
Taxi &   \begin{tabular}[c]{@{}l@{}} 
    \textbf{Context: }\\
    \textit{I wager that is majorly low. All the \underline{\textbf{taxi}} drivers around me drive} \\ 
    \textit{brand new hybrid \underline{\textbf{Lexus}}’s. If you consider the fuel, cars upkeep,}\\ 
    \textit{the car itself and the insurance. They must be owning a good}\\ 
    \textit{scoop to make all that worth it.}\\
   \end{tabular} & \begin{tabular}[c]{@{}l@{}}
    \textbf{True Response: } \\
    \textit{A lot of \underline{\textbf{taxi}} drivers round my way are working two jobs and have it as their}\\
    \textit{second gig filling in what little free time they have from their main job.}\vspace{1.4mm}\\ 
    \textbf{False Response: }\\
    \textit{Buying vehicles (converged ones in my fathers case) is a huge expense which}\\
    \textit{I don’t think can be fully tax offset.}\\
   \end{tabular} \\
  \midrule
Restaurant &   \begin{tabular}[c]{@{}l@{}} \\
\textbf{Context: }\\ 
    \textit{Interesting. Thanks for the post and thanks for mentioning}\\
    \textit{Normandie. I will definitely check that out and look at staying}\\
    \textit{somewhere other than Zocalo. Any other recommendations}\\
    \textit{for stuff you really liked? I'm a huge food guy so any awesome}\\
    \textit{\underline{\textbf{restaurants}} (already have a Pujol \underline{\textbf{reservation}}) are welcome.} \\
   \end{tabular} & \begin{tabular}[c]{@{}l@{}}
    \textbf{True Response: } \\
    \textit{You're welcome. Thanks for reading. Don't get me wrong Zocalo has some}\\
    \textit{historic significance etc. and is nice to visit for the day, but that's about all}\\
    \textit{the time you need there. For some \underline{\textbf{cheap}} but still good tacos, ...}\vspace{1.4mm}\\ 
    \textbf{False Response: }\\
    \textit{Zocalo is hectic and filled with tons of people. IMO after 1 day there you'll}\\
    \textit{want out. Roma Norte and Condesa have some beautiful parks and are filled}\\
    \textit{with cool cafes, restaurants and bars ...}\\
   \end{tabular} \\
\midrule
Train &  \begin{tabular}[c]{@{}l@{}}
\textbf{Context: }\\ 
    \textit{You just need to hope you don't need to walk all the way to the} \\
    \textit{back of the \underline{\textbf{train}}.}\\
   \end{tabular} & \begin{tabular}[c]{@{}l@{}}
    \textbf{True Response: } \\
    \textit{I have to do I multiple times a day with the TGV's. Those are only 200m short.}\\
    \textit{I don't working on this 400m \underline{\textbf{train}} often. But yes it happens.}\vspace{1.4mm}\\ 
    \textbf{False Response: }\\
    \textit{We need a sub for European trains!}\\
   \end{tabular} \\
\midrule
Hotel &   \begin{tabular}[c]{@{}l@{}} 
\textbf{Context: }\\ 
    \textit{Thanks for the info. I didn't book a \underline{\textbf{hotel}} yet and plan to do that}\\
    \textit{by tomorrow. Wasn't aware that most don't have \underline{\textbf{free parking}}.}\\
    \textit{I'll try to find one with \underline{\textbf{parking}} included.} \\
   \end{tabular} & \begin{tabular}[c]{@{}l@{}}
    \textbf{True Response: } \\
    \textit{You are not likely to find a \underline{\textbf{hotel}} with \underline{\textbf{free parking}} in the old city. And, to be}\\
    \textit{honest, unless budget is a big deal, for a short trip it's entirely worth the}\\
    \textit{experience to stay in either the upper or lower old city ...}\vspace{1.4mm}\\ 
    \textbf{False Response: }\\
    \textit{Where is your hotel? Many either have parking, or arrangements for parking}\\
    \textit{in nearby lots and garages. If you're at or near the Frontenac there is a public}\\
    \textit{garage under city hall that is much less expensive than many hotel options.}\\
   \end{tabular} \\
\midrule
Attraction &   \begin{tabular}[c]{@{}l@{}} 
\textbf{Context: }\\ 
    \textit{Thank youuu! I'll better pack a coat to keep myself warm! Hmm}\\
    \textit{you’re right I might just skip the day trip! I like history/\underline{\textbf{museum}},}\\
    \textit{art, \underline{\textbf{architecture}} and scenery/nature! What are the top few places}\\
    \textit{do you recommend though?} \\
   \end{tabular} & \begin{tabular}[c]{@{}l@{}}
    \textbf{True Response: } \\
    \textit{In terms of \underline{\textbf{museums}} and history, you're really spoilt for choice in London.}\\
    \textit{The Natural History \underline{\textbf{Museum}}, Imperial War \underline{\textbf{Museum}} and National Maritime}\\
    \textit{\underline{\textbf{Museum}} are my personal favourites. If you like nature go check out the}\\
    \textit{wildlife in Richmond \underline{\textbf{Park}}. It's a ...}\vspace{1.4mm}\\ 
    \textbf{False Response: }\\
    \textit{The UK is due to be extremely cold this winter so I’d have some extra warm}\\
    \textit{clothes just in case. November is usually fine, a bit rainy, but this year might}\\
    \textit{be a special case. You can visit Camden but I personally wouldn't spend ...}\\
   \end{tabular} \\
\bottomrule
\end{tabular}%
}
\caption{\label{tab:domainreddit} Example from \textsc{DomainReddit} dataset, where the salient domain terms are marked as \underline{\textbf{bold}}. The texts are displayed in the original version, without correcting typos.}

\end{table}


\end{document}